\documentclass[journal]{IEEEtai}

\usepackage[colorlinks,urlcolor=blue,linkcolor=blue,citecolor=blue]{hyperref}

\usepackage{color,array}
\usepackage{graphicx}
\usepackage{booktabs}
\usepackage{amsmath}
\usepackage{multirow}
\usepackage[numbers]{natbib}

\begin{document}

\title{A Multimodal Adaptive Graph-based Intelligent Classification Model for Fake News}

\author{Junhao (Leo) Xu

\thanks{Junhao (Leo) Xu is with Guangzhou College of Technology and Business, Guangzhou, China (e-mail: xujunhao1105@gmail.com).}}

% \author{}
% % \markboth{Journal of IEEE Transactions on Artificial Intelligence, Vol. 00, No. 0, Month 2020}
% % {First A. Author \MakeLowercase{\textit{et al.}}: Bare Demo of IEEEtai.cls for IEEE Journals of IEEE Transactions on Artificial Intelligence}

\maketitle

\begin{abstract}
Numerous studies have been proposed to detect fake news focusing on multi-modalities based on machine and/or deep learning. However, studies focusing on graph-based structures using geometric deep learning are lacking. To address this challenge, we introduce the Multimodal Adaptive Graph-based Intelligent Classification (aptly referred to as MAGIC) for fake news detection. Specifically, the  Encoder Representations from Transformers was used for text vectorization whilst ResNet50 was used for images. A comprehensive information interaction graph was built using the adaptive Graph Attention Network before classifying the multimodal input through the Softmax function. MAGIC was trained and tested on two fake news datasets, that is, Fakeddit (English) and Multimodal Fake News Detection (Chinese), with the model achieving an accuracy of 98.8\% and 86.3\%, respectively. Ablation experiments also revealed MAGIC to yield superior performance across both the datasets. Findings show that a graph-based deep learning adaptive model is effective in detecting multimodal fake news, surpassing state-of-the-art methods.
\end{abstract}


\section{Introduction}
The rapid evolution of Online Social Networks (OSNs) has revolutionized traditional communication methods, enabling content generation and dissemination to many users within mere seconds. As a result, OSNs have emerged as the predominant platforms for daily information sharing \cite{1,2}. However, as society relishes the conveniences of OSNs, the lag in robust network supervision and the absence of advanced technologies have paved the way for fake news to thrive \cite{3}. Fake news, defined as any misleading, false, or incorrect information that is deliberately propagated, encompasses rumors, hoaxes, and conspiracy theories while relying on the dissemination of deceptive, unverified, and intentionally misleading content, thus posing an increasingly intricate challenge to societal norms and our everyday activities \cite{4}.

Research in fake news detection has experienced substantial advancements due to machine and deep learning and can be categorized into two types: modality-based and graph structure-based. Modality-based studies capture features such as semantic \cite{5}, emotion \cite{6,7}, stance-based features \cite{8}, intent \cite{9}, and user profiles \cite{10} from textual and visual input, and fuse them together using the conventional machine and/or deep learning algorithms \cite{11,12,13,14}. 

On the other hand, graph structure-based studies utilizing deep learning algorithms (i.e., Graph Neural Networks (GNN)) have shown remarkable success in natural language processing \cite{15,16}, including detecting fake news on OSNs which can be attributed to the exponential growth of users, news content, and user interactions within these networks \cite{10,17}. OSNs evolve into complex graph structures \cite{18,19}, hence posing challenges for the conventional machine and deep learning models \cite{20}. 

Studies integrating multimodal input with graph structure-based mechanisms are lacking. In fact, to the best of our knowledge only the work of Bi and colleagues \cite{21} was found in which the authors proposed a heterogeneous attention network and tested their model using images and text from Weibo. To fill this gap, we introduce the Multimodal Adaptive Graph-based Intelligent Classification (herein referred to as MAGIC) - a model that aims to detect fake news through textual (i.e., original posts as well as their accompanying comments) and visual contents using an adaptive residual deep-focus Graph Attention Network (GAN). Experiments with two datasets revealed our model to be superior. 

The remainder of this paper is organized as follows: Section \ref{s1} delves into a comprehensive review of the related work, followed by the methodology in Section \ref{s3}. Section \ref{s4} introduces our experimental details and evaluation methods. A detailed presentation of our experimental results and discussion is provided in Section \ref{s5}. Finally, Section \ref{s6} offers a summary of the entire paper, outlining its limitations and future directions.

\section{Related work}\label{s1}
This section describes previous studies on fake news detection, highlighting the key gaps related to this study.
\subsection{Single-modal fake news detection}
In the early stages of fake news detection, numerous methodologies were deployed to gauge the veracity of information based solely on its textual content. Predominantly, these techniques revolved around the extraction of textual features at statistical or semantic levels \cite{1,22,23,24}. Most of the studies have utilized deep learning algorithms, for example, the Recurrent Neural Network-based Gatekeeper Behavior Model was proposed to detect fake news in OSNs, leveraging RNN’s capacity to capture evolving contextual information from pertinent posts \cite{25} whilst Ma and peers \cite{26} developed a Dual-Channel Convolutional Neural Network (CNN) to detect fake news with attention pooling. On the other hand, Qi and peers \cite{27} designed a CNN model to autonomously detect false content within images with promising results. 

\subsection{Multimodal fake news detection}
With the emergence of multimodal forms of information on OSNs including text, videos, audios and images, researchers began to explore mechanisms to detect fake news using more than a single form of input \cite{28,29,30}. For instance, a hybrid deep learning model was introduced to detect fake news based on text and images, whereby the authors proposed an Attention-based Multimodal Factorized Bilinear Pooling mechanism to maximize the correlation between the textual and visual features. It achieved an accuracy of 88.3\% on a Twitter dataset \cite{29}. Giachanou and peers \cite{31} used the Visual Geometry Group Network (VGGNet) and the Bidirectional Encoder Representations from Transformers (BERT) to compute similarities between images and article headlines, and thus enhancing the effectiveness of fake news detection.
Similarly, Qi and peers \cite{30} proposed a novel entity-enhanced multimodal fusion framework, which simultaneously models cross-modal correlations to detect diverse multimodal fake news, achieving an accuracy of 90.4\% on Weibo (i.e., Chinese dataset) and 97.5\% on Politifact (i.e., English tweets). Finally, Jing and peers \cite{28} introduced a Multimodal Progressive Fusion Network (MPFN) that captures representations at various levels for text and images and establishes strong connections between them. The authors reported an accuracy of 83.3\% on a Twitter dataset.

\subsection{Graph-structured fake news detection}
Studies focusing on the graph structured mechanism to detect fake news are limited. For example, a geometric deep learning (GDL) model based on text and the structure of social media conversations was introduced for fake news detection by Tian and peers who proposed the Detection with User and Comment Networks (DUCK) \cite{17}. DUCK underscores the role of comments in capturing user reactions to information, hence demonstrating that comment structures provide supplementary signals for detecting fake news. Yang and peers \cite{16} introduced an innovative model known as the Graph Attention Capsule Network (GACN) on dynamic propagation structures that segments the fake news propagation structure, chronologically into multiple static graphs to capture the dynamic interaction features. GACN was reported to achieve accuracy rates of 88.9\% and 90.0\% on the Twitter15 and Twitter16 datasets, respectively.
Studies integrating multimodal input with graph structure-based mechanisms are lacking. Only a single study was found in which the authors proposed a heterogeneous graph model that incorporates texts and images for fake news detection \cite{21}. The authors modelled the information propagation network of Weibo as a heterogeneous graph containing various semantic information and achieved accuracy scores exceeding 92\% on the Weibo2016 and Weibo2021 datasets.

\section{Methodology}\label{s3}
\begin{figure*}[t]
    \centering
    \includegraphics[width=1\linewidth]{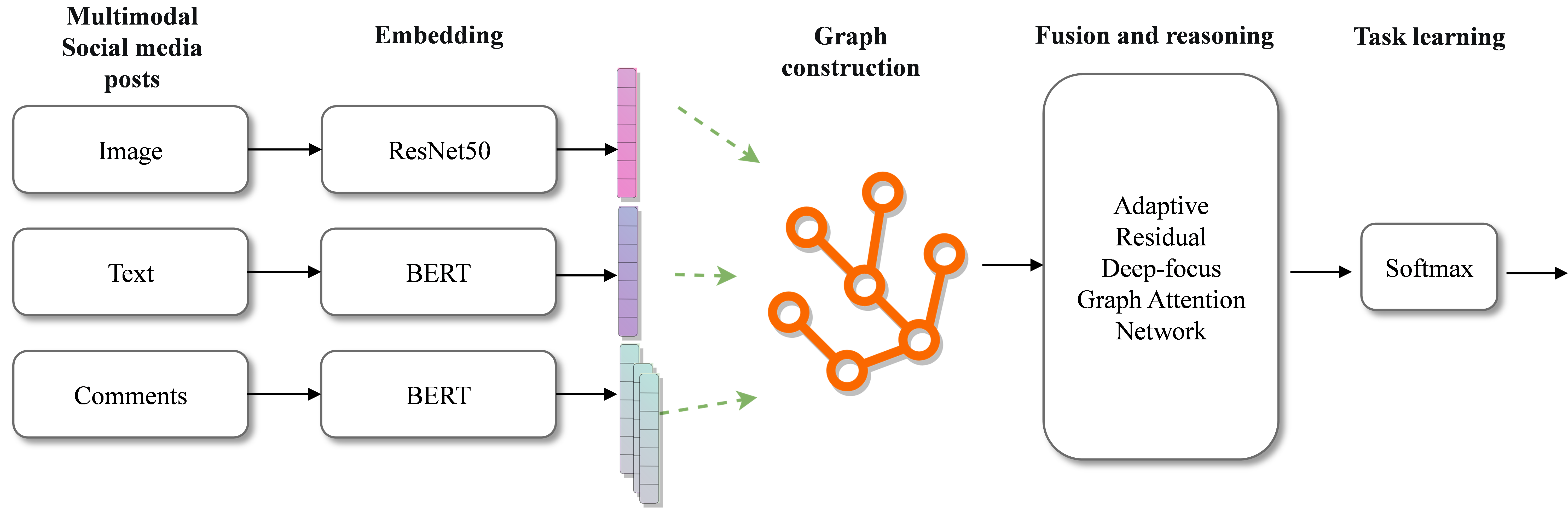}
    \caption{Architecture of Multimodal Adaptive Graph-based
Intelligent Classification Model (MAGIC).}
    \label{fig:enter-label}
\end{figure*}
Figure \ref{fig:enter-label} depicts the main pipeline of MAGIC encompassing four key stages: Multimodal social media posts (dataset), Embedding, Graph construction, Fusion and reasoning, and ultimately, Task learning.
\subsection{Multimodal social media posts}
Data will be collected from social media posts containing multimodal information such as text, image, and Comment relations, etc. Two fake news datasets in English and Chinese were sourced to train and test our proposed model. They are:

- \textbf{Fakeddit}\footnote{https://github.com/entitize/Fakeddit}: an English multimodal dataset containing text, comments, and images from Reddit (n = 3,127). The samples within this dataset are labeled using a 2-way (real, fake) and a 3-way classification (real, fake with true text, fake with false text) scheme. Modelling was performed using both these classifications.

- \textbf{Multimodal Fake News Detection dataset}\footnote{https://data.beijing.gov.cn/kjzy2020/index.html}(MFND): a multimodal Chinese dataset containing text, comments, and images from Weibo (n = 2,953). The dataset is categorized into three distinct classes: uncertain, fake news, and real news.

The number of instances for each classified labels in the datasets are given in Table \ref{tab:dataset_overview}, indicating the datasets to be balanced.
\begin{table}[ht]
    \centering
    \caption{Fake news datasets overview}
    \begin{tabular}{c|c|c|c|c}
        \toprule
        Dataset & 2-way & 3-way & n & Total \\
        \midrule
         & - & real & 1,000 &  \\
        MFND& - & fake & 953 & 2,953\\
        & & uncertain & 1,000 & \\
        \midrule
         & real & real & 1,048 &  \\
        Fakeddit& fake & fake with true text & 1,060 & 3,127\\
        & & fake with false text & 1,019 & \\
        \bottomrule
    \end{tabular}
    \label{tab:dataset_overview}
\end{table}

Observing the two datasets, it becomes evident that the distribution of categories is relatively balanced, with no significant gaps between individual labels. This balanced distribution provides us with a more reliable and comprehensive foundation for accurate analysis and modeling. In the fake news dataset, each piece of news is composed of different modalities, namely posts, images, and comments. The following Figure \ref{fig:2} illustrates the overall distribution of these modalities.
\begin{figure}[h]
    \centering
    \includegraphics[width=\linewidth]{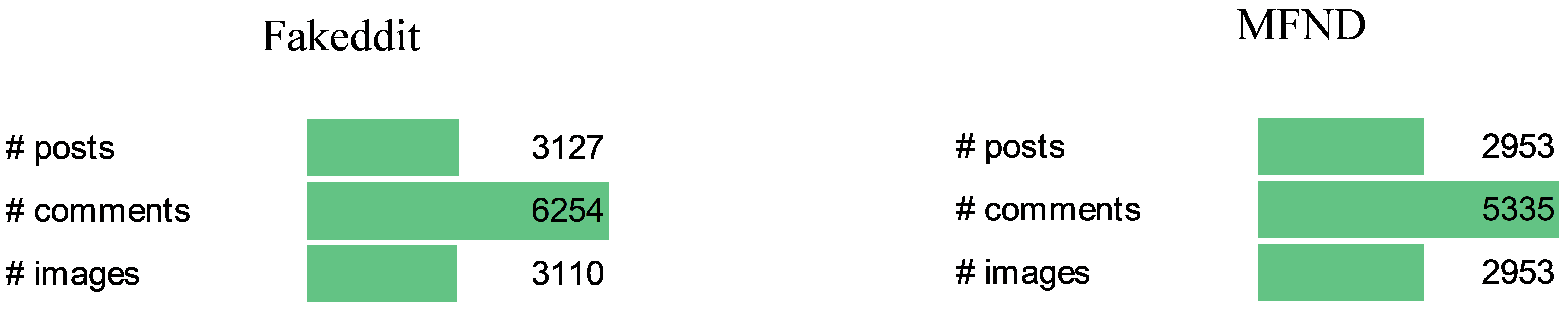}
    \caption{Distribution of modalities}
    \label{fig:2}
\end{figure}

\begin{figure}[h]
    \centering
    \includegraphics[width=\linewidth]{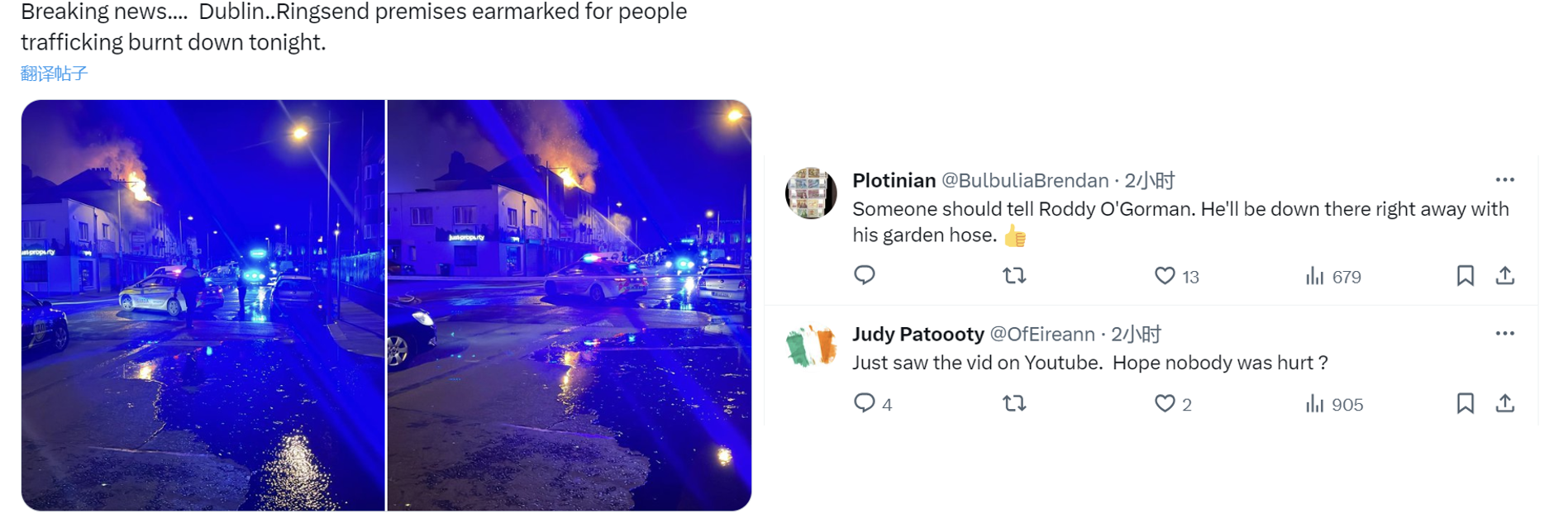}
    \caption{Example of dataset}
    \label{fig:3}
\end{figure}

Taking a sample from the dataset, as illustrated in Figure \ref{fig:3}, we observe a user posting 
a narrative along with an image, while numerous other users engage in comments. The 
semantic gap between these comments and the post is substantial, mostly comprising
emotional opinions. Some users even verify the authenticity of the information, as evident 
in this example where a commenter claims to have seen related information on YouTube. 
Thus, we affirm the dataset's multimodal and semantic diversity.

\subsection{Embedding}
Fake news exhibit diverse modalities; thus, the study considered text (i.e., text and their 
respective comments) and images, hence two forms of embeddings were done, as 
follows:
\subsubsection{Textual embedding}
\begin{figure}[h]
    \centering
    \includegraphics[width=1\linewidth]{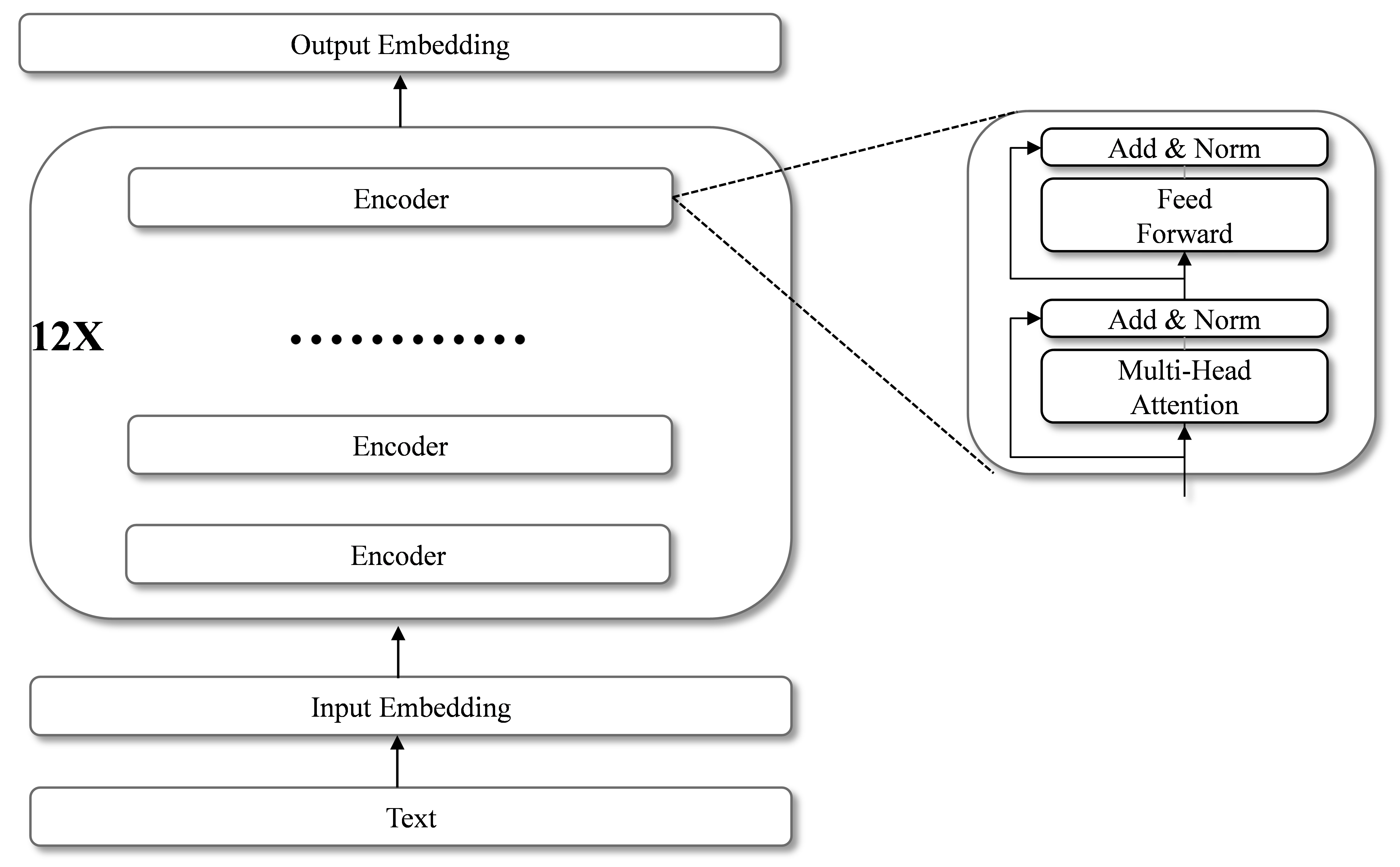}
    \caption{BERT-base architecture}
    \label{fig:4}
\end{figure}

BERT, a well-known transformer-based encoder that supports multiple languages including English and Chinese was used to convert the textual input into its vector representations \cite{32}. Specifically, the BERT-base variant containing 12 transformer encoder blocks was used to extract the vector representations, whereby these encoders convert each tokenized word into its corresponding numerical vector, effectively mapping semantically related words to numerically proximate embeddings (see Figure \ref{fig:4}).
We loaded a pre-trained BERT model, inputted the text into the model, and set the 
final output embedding to 768 dimensions. The obtained embedding vector serves as a 
representation of the semantic encoding for the entire sentence:
\begin{equation}
    S_n=BRET([w_n^0,\quad w_n^1,...,w_n^m]),
\end{equation}
where m symbolizes the length of the word sequence, while n stands as an index delineating the ordinal position of the sentence in consideration. The 768-dimensional feature vector corresponding to the $m^{th}$ word within the $n^{th}$ sentence is represented by $w_n^m$. Furthermore, $S_n$ designates the feature vector of the $n^{th}$ sentence, as encoded by the BERT pre-trained model.
\subsubsection{Image embedding}
ResNet50, which is a residual network was used to transform the images into their vector representations (Figure \ref{fig:5}). ResNet50 is a 50-layer network architecture that uses the residual connections allowing information to be skipped over one or more layers in the network, hence allowing deep networks to extract low-level, mid-level, and high-level features from images while adapting to their depth \cite{34}.
\begin{figure}
    \centering
    \includegraphics[width=1\linewidth]{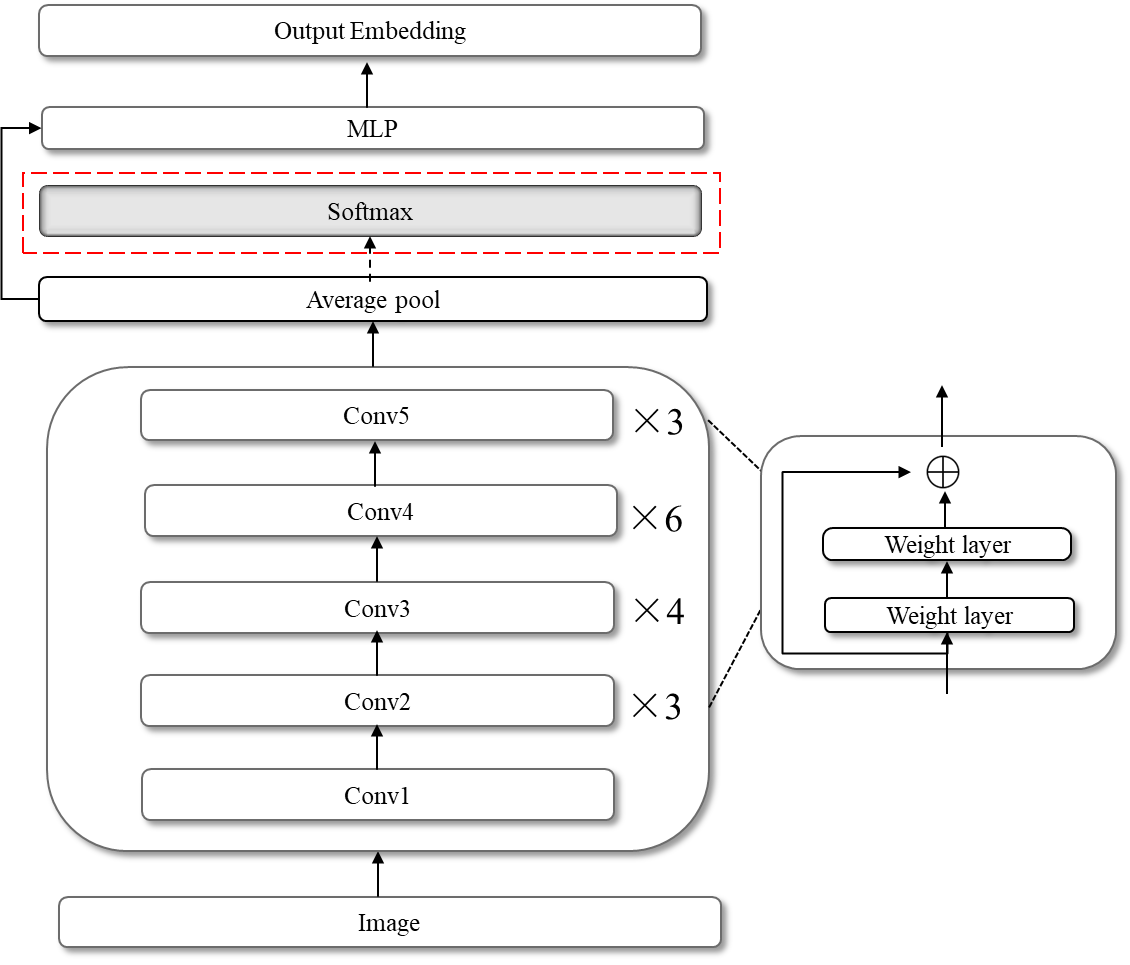}
    \caption{ResNet50 architecture}
    \label{fig:5}
\end{figure}
We loaded a pre-trained ResNet50 model, removed the Softmax layer, and instead 
added a Multiple Layer Perceptron (MLP) layer to map the extracted image vector 
features to the same dimension (i.e., 768 dimension) as the text features mentioned above. 
The image is eventually represented as:
\begin{equation}
    V_n=ReLU(MLP(ResNet50(x_n)),
\end{equation}
where $x_n$ designates the $n^{th}$ image. Relu represents the rectified linear unit, an activation function ubiquitously utilized in neural architectures. Post extraction of salient features from the image via the ResNet-50 paradigm, the combined operations of MLP and Relu ensure the alignment of these features into a vectorial space that shares the same dimensionality with that of the textual vector, denoted herein as V.
\subsection{Graph construction}
As shown in Figure \ref{fig:enter-label}, the output of these embedding techniques is then fed into the graph construction phase, in which each sample is formulated into an information graph pertaining to its associated message as shown in Figure \ref{fig:5}.
\begin{figure}
    \centering
    \includegraphics[width=1\linewidth]{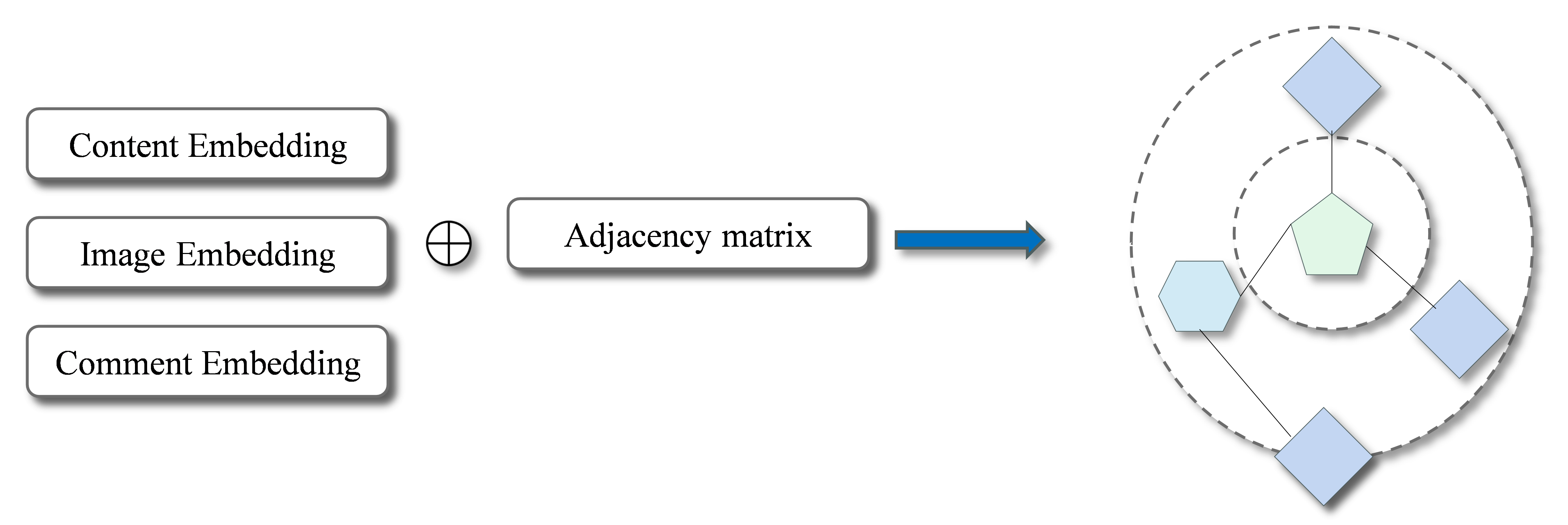}
    \caption{Graph construction}
    \label{fig:5}
\end{figure}
Specifically, the adjacency matrix was used to define the connected edges based on 
the vector representations for the textual (including content of the posts and comments) 
and visual content (different colors in the graph represent different types of vector 
representations). This results in the creation of the multimodal interaction graph, which 
captures the essence of interactions among these three elements. The mathematical 
formula is defined as follows:
\begin{equation}\label{eq3}
    G=(P,V,C,A),
\end{equation}
where P stands for the embedded representation of the content of the posts, while V signifies the embedded depiction of the associated image. C designates the embedded comments furnished by participants in response to the posts. The adjacency matrix is represented by A, encapsulating the structural relationships among these entities. The multimodal interaction graph, capturing the essence of interplays among these elements, is denoted as G.
\subsection{Fusion and reasoning}
\begin{figure*}
    \centering
    \includegraphics[width=1\linewidth]{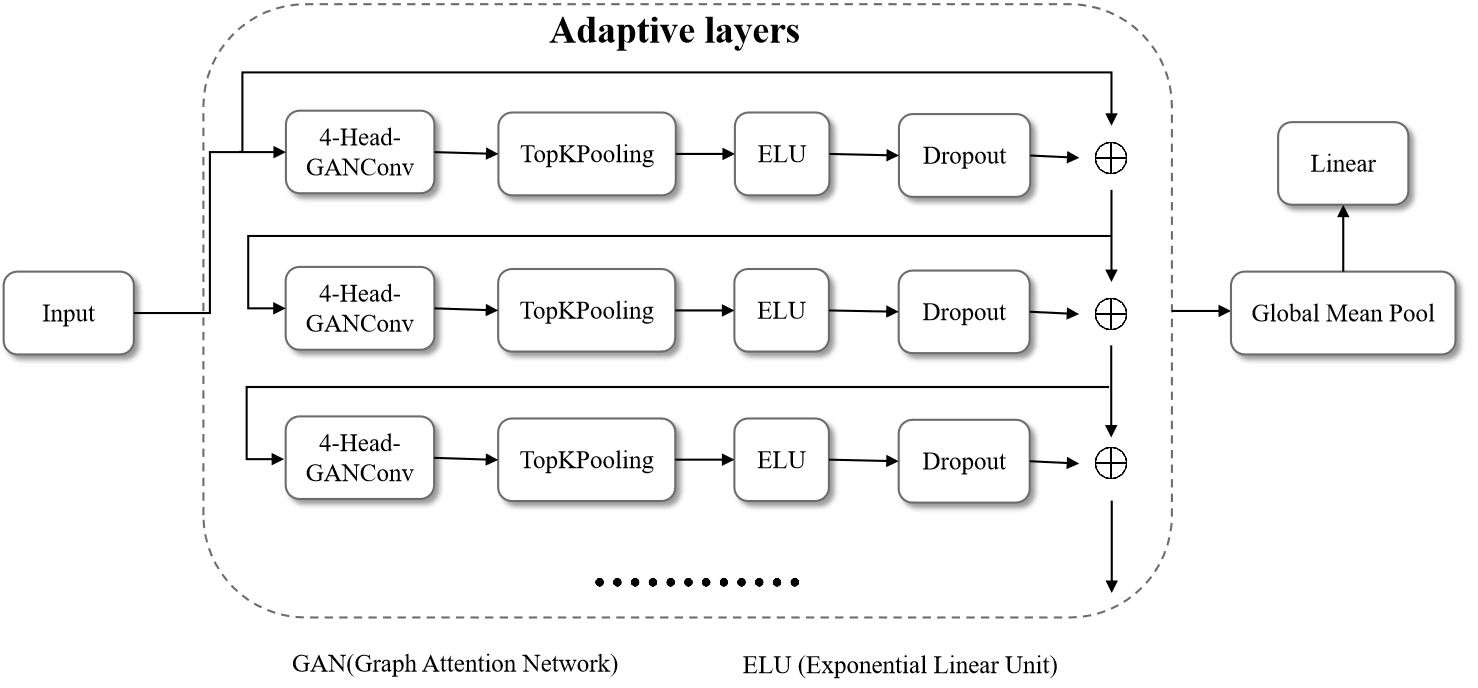}
    \caption{Adaptive Residual Deep-focus Graph Attention Network 
architecture}
    \label{fig:6}
\end{figure*}

Upon obtaining the multimodal graph defined by Formula \ref{eq3}, it is imperative to 
integrate and infer from it using a graph neural network. The adaptive residual network 
was adapted to fuse the multimodal input as shown in Figure \ref{fig:6}. The number of layers 
was set as a variable parameter, allowing the model to automatically search within a 
specified range of layers and save the results with the optimal number of layers.
This network architecture establishes residual graph connections and the formulation 
for this process is delineated by the following equation:

\begin{equation}
\begin{aligned}
& \begin{cases}
G(x)_1 = F(x) \\
G(x)_n = G(x)_{n-1} + F(G(x)_{n-1})
\end{cases}
\end{aligned}
\end{equation}
where $n$ denotes the number of residual layers, $G(x)_n$ signifies the feature representation of the entire graph after $n$ layers, and $F$ represents the mapping function of
the residual block \cite{34}.
The Graph Attention Network, which is a variant of GNNs, plays a critical role in 
enhancing graph-based data processing by employing an attention mechanism to 
aggregate node features and adaptively extract contextually relevant information 
\cite{35}. The attention coefficients are first calculated through a multi-head attention mechanism to represent the similarity scores between the central node and 
its neighboring node. These coefficients are used to update the information of the 
neighboring nodes as well as the node itself. This adaptive attention mechanism 
contributes to the comprehension of intricate modality relationships and structural 
patterns within graphs, as defined below:
\begin{equation}
    e_{ij}=a(\vec{h}_i,\vec{h}_j),
\end{equation}
\begin{equation}
    \alpha_{ij}=\frac{\exp(e_{ij})}{\sum_{k\in\mathcal{N}_i}\exp(e_{ik})},
\end{equation}
where $h$ represents node features, $\alpha$ is a parameter vector subject to learning, and $e$, denotes unnormalized coefficients for pairs of nodes $i$ and $j$,based on their respective features. The coefficient $\alpha_{ij}$ signifies the attention weight assigned between nodes $i$ and $j$,thereby reflecting the influence of graph structure. This attention mechanism restricts the focus of node i to its local neighborhood, denoted by $j\in\mathcal{N}_i.$

In order to enhance the stability of the self-attention learning process, the utilization of 
multi-head attention proves to be advantageous. This entails 
independently replicating the operations of this layer K times (each copy possessing 
distinct parameters), and subsequently aggregating the outputs through feature 
concatenation or addition:
\begin{equation}
\vec{h}_i'=||_{k=1}^K\sigma(\sum_{j\in\mathcal{N}_i}\alpha_{ij}^kW^k\vec{h}_j),
\end{equation}
where $a^k$ signifies the attention coefficients computed by the k-th iteration, while W$^{k}$ denotes the weight matrix that delineates the linear transformation inherent to the k-th iteration. $\mathcal{N}_i$ represents the set of neighboring nodes for node i. Here, $\sigma$ represents an activation function, such as the Leaky ReLU. $\vec{h}_i^{\prime}$ denotes the features after aggregating updates from neighboring nodes and the node itself.

In the context of fake mews detection, not all textual comments and images are equally significant, hence the model further refines and consolidates the most crucial and valuable node information by utilizing the top-k pooling mechanism. Mathematically, for each node $i$,the top-k pooling operation involves selecting the k highest attention coefficients umong its neighbors $j\in\mathcal{N}_i$ and redistributing the remaining attention weights proportionally. This can be represented as:
\begin{equation}
\alpha_{ij} = 
\begin{cases} 
\alpha_{ij}, & 
\begin{array}{l}
\text{if } \alpha_{ij} \text{ is among the k highest} \text{coefficients} \\
\text{for node  i},
\end{array} \\
0, & \text{otherwise.}
\end{cases}
\end{equation}

This weight updating mechanism serves to facilitate a secondary focus on more valuable node information during the update process. This refinement aids in directing attention towards nodes that carry greater significance, thereby enhancing the model's ability to discern and prioritize crucial features. In this study, we retained the top $80\%$ of nodes for the pooling, which then goes through the Exponential Linear Unit (ELU) that enhances the model's performance and generalization capability. A dropout rate of 0.20 was used to mitigate overfitting within our model.

The output from last layer is aggregated using the Global Mean Pool to create a fixedlength representation of the entire graph's features. Given a graph with nodes V and their associated feature vectors $x_v$ for each node v$\in\mathcal{V}$, the Global Mean Pooling operation computes the mean of all node feature vectors across the entire multimodal information graph. The mathematical mechanism of Global Mean Pooling (GMP) can be described as
follows:
\begin{equation}
GMP(X)=\frac1{|V|}{\sum_{v\in V}X_v},
\end{equation}
where GMP(X) represents the global mean pooled feature vector obtained from the node feature vectors $x_{v}$, and $|V|$ signifies the total number of nodes in the graph Subsequently, this pooled feature representation can be utilized for downstream tasks, such as graph classification for fake news detection.
\subsection{Task modelling}
Finally, Softmax was used to classify the fake news classification. This step involves utilizing a loss function that steers the model towards minimizing the cross-entropy erron associated with a specific training instance, considering the ground-truth label $y$
\begin{equation}
p=softmax(W_cX+b_c),
\end{equation}
\begin{equation}
L=-\sum y{\log p},
\end{equation}
where $W_c$ and $b_c$ represent the parameters in the fully connected layer (utilized for mapping the final dimensions to the category vector), while p represents the category probabilities obtained after applying the softmax function.The datasets were partitioned into 80-20 (i.e., training-testing). Within the training set, a further division is made into 80\% actual training data and 20\% validation data. The 80\% actual training data is employed for the iterative parameter updates through backpropagation, while model evaluation is conducted on the 20\% validation data. Ultimately, the set of model parameters that performs best on the validation set is saved for the final evaluation on the test set.
\section{Experiments and evaluations}\label{s4}
In this section, we will elaborate on the specific equipment and software employed in 
the experiments. Additionally, we will delve into the detailed configuration of training 
parameters for our model, compare it with the chosen baseline, and present the evaluation 
metrics employed.
\subsection{Experimental setups}
All the training and testing were done using Python, utilizing various libraries 
available such as sklearn, pytorch and Pyg, with a GPU NVIDIA Tesla P100. Learning 
rate and batch size are set to 0.002 and 128, respectively. To optimize model parameters 
effectively, the Adam optimizer is also utilized. We experimented several models as 
outlined below:

-BERT \cite{39}: Only textual original posts were used as input 
for fake news classification. In this study, researchers achieved remarkable results in 
detecting fake news by fine-tuning the pre-trained BERT model.

-BERT+CNN \cite{40}: The model takes into consideration all the 
features encoded by BERT for sentences and feeds the resultant output into a CNN layer, 
effectively enabling the detection of fake news in textual content.

-BERT+LSTM \cite{41}: By incorporating BERT's comprehensive 
feature extractions and funneling the output through an LSTM layer, this architecture 
significantly boosts the model's capacity to identify fake news by capturing sequential 
patterns and contextual intricacies in the text.

-BERT+BiLSTM \cite{40}: This model leverages the entirety of 
features encoded by BERT for sentence comprehension and subsequently directs the 
resulting output to a Bidirectional Long Short-Term Memory (BiLSTM) layer. This 
design is aimed at effectively capturing temporal dependencies and contextual nuances 
within the text, thereby enhancing the capability to detect fake news in the given textual 
content.

-BERT+CNN+BiLSTM \cite{42}: This model utilizes CNN-BiLSTM to 
capture the local, global, and temporal meanings of sentences, achieving excellent 
performance.

-ResNet50 \cite{43}: Only the images were used as input for fake news classification. Utilizing pre-trained ResNet for enhanced fake news image representation has proven to be effective in 
capturing meaningful features from images associated with disinformation.

-BERT-GNN \cite{44}: Graph-based mechanism using the textual input (original post and 
comments).

It is to note that a combination of ResNet and GNN is not feasible as each news article 
only has one or no image (i.e., one node, no image encoded as an all-0 feature node), 
hence it is not possible to form a graph. All the models were trained and tested on both 
the datasets, to assess their robustness in handling different languages. The experiments 
above were conducted to assess the model’s performance based on single modal input, 
using the conventional approach. These act as our baseline models.

We also conducted ablation experiments to validate the effectiveness of each module 
within our model. This involved systematically analyzing and selectively removing 
components to assess their individual contributions to the overall performance.

\subsection{Evaluation}
As this is a classification problem, the standard metrics were used to assess the 
effectiveness of MAGIC. The performance of the model can be illustrated using a 
confusion matrix, as depicted in Table \ref{tab3}.
\begin{table}[h!]
\centering
\label{tab3}
\caption{Confusion Matrix}
\begin{tabular}{|c|c|c|c|}
\hline
\multicolumn{2}{|c|}{\textbf{}} & \multicolumn{2}{c|}{\textbf{Predicted}} \\ \hline
\multicolumn{2}{|c|}{\textbf{}} & \textbf{Positive} & \textbf{Negative} \\ \hline
\multirow{2}{*}{\textbf{Actual}} & \textbf{Positive} & True Positive (TP) & False Negative (FN) \\ \cline{2-4} 
 & \textbf{Negative} & False Positive (FP) & True Negative (TN) \\ \hline
\end{tabular}
\label{tab3}
\end{table}
From the confusion matrix, metrics like accuracy, precision, recall and F1-score can 
be computed.

Accuracy measures the proportion of correctly classified instances among all instances.
\begin{equation}
Acc=\frac{(True\:Positives\:+\:True\:Negatives)}{Total\:Samples}.
\end{equation}

Precision represents the proportion of correctly predicted positive instances out of all
instances predicted as positive.

\begin{equation}
Pre=\frac{TruePositives}{(TruePositives+FalsePositives)}.
\end{equation}

Recall indicates the proportion of correctly predicted positive instances out of all actual positive instances.

\begin{equation}
Rec=\frac{True\:Positives}{(True\:Positives\:+\:False\:Negatives)}.
\end{equation}

F1-score combines precision and recall into a single value, providing a balanced
measure of a model's accuracy in identifying both positive and negative instances

\begin{equation}
F1=2*\frac{(Precision*Recall)}{(Precision+Recall)}.
\end{equation}
In this study, the aforementioned four indicators are reported in the performance 
evaluation. Our MAGIC model is compared to baseline models reported in other relevant 
studies and, similarly, conducts ablation experiments to validate the effectiveness of its 
modules. To ensure a fair comparison with other models, uniform training strategies and 
parameters are employed in this research.

\section{Results and discussion}\label{s5}
The overall results for the fake news classification for MAGIC are presented in this section, followed by the ablation experiments. Table \ref{tab4} illustrates the performance comparison between MAGIC and the baseline models across the two datasets, whilst Table \ref{tab5} provides the confusion matrix.

\begin{table*}[h!]
\centering
\caption{ Performance of MAGIC on two datasets (in \%). Items in bold indicate best results.}
\begin{tabular}{c|c|c|c|c|c|c}
\hline
\textbf{Dataset} & \textbf{Modality} & \textbf{Method} & \textbf{Acc} & \textbf{Pre} & \textbf{Rec} & \textbf{F-1} \\ \hline
\multirow{8}{*}{Fakeddit (2-way)} & Text & BERT & 94.41 & 95.09 & 92.31 & 93.51 \\ \cline{2-7}
 & Text & BERT+CNN & 96.33 & 96.50 & 95.20 & 95.80 \\ \cline{2-7}
 & Text & BERT+LSTM & 95.05 & 95.55 & 93.27 & 94.28 \\ \cline{2-7}
 & Text & BERT+BiLSTM & 95.21 & 93.94 & 95.69 & 94.71 \\ \cline{2-7}
 & Text & BERT+CNN+BiLSTM & 95.53 & 95.90 & 93.99 & 94.86 \\ \cline{2-7}
 & Text & BERT-GNN & 95.53 & 94.92 & 95.12 & 95.02 \\ \cline{2-7}
 & Image & ResNet50 & 94.89 & 93.94 & 95.01 & 94.43 \\ \cline{2-7}
 & Text+Image & MAGIC (ours) & \textbf{98.72} & \textbf{98.68} & \textbf{98.44} & \textbf{98.56} \\ \hline
\multirow{8}{*}{Fakeddit (3-way)} & Text & BERT & 94.41 & 94.64 & 94.38 & 94.35 \\ \cline{2-7}
 & Text & BERT+CNN & 94.09 & 94.21 & 94.05 & 94.02 \\ \cline{2-7}
 & Text & BERT+LSTM & 95.05 & 95.05 & 95.03 & 95.03 \\ \cline{2-7}
 & Text & BERT+BiLSTM & 94.25 & 94.49 & 94.22 & 94.19 \\ \cline{2-7}
 & Text & BERT+CNN+BiLSTM & 93.29 & 93.71 & 93.25 & 93.19 \\ \cline{2-7}
 & Text & BERT-GNN & 94.57 & 94.26 & 94.35 & 94.27 \\ \cline{2-7}
 & Image & ResNet50 & 94.41 & 94.83 & 94.62 & 94.61 \\ \cline{2-7}
 & Text+Image & MAGIC (ours) & \textbf{97.60} & \textbf{97.60} & \textbf{97.60} & \textbf{97.60} \\ \hline
\multirow{8}{*}{MFND (3-way)} & Text & BERT & 77.33 & 79.26 & 77.72 & 77.35 \\ \cline{2-7}
 & Text & BERT+CNN & 77.83 & 78.72 & 78.23 & 77.51 \\ \cline{2-7}
 & Text & BERT+LSTM & 81.22 & 81.86 & 81.20 & 81.30 \\ \cline{2-7}
 & Text & BERT+BiLSTM & 81.22 & 81.27 & 81.32 & 81.10 \\ \cline{2-7}
 & Text & BERT+CNN+BiLSTM & 80.20 & 81.05 & 80.19 & 80.16 \\ \cline{2-7}
 & Text & BERT-GNN & 83.42 & 83.71 & 83.71 & 83.71 \\ \cline{2-7}
 & Image & ResNet50 & 52.62 & 54.49 & 52.83 & 48.84 \\ \cline{2-7}
 & Text+Image & MAGIC (ours) & \textbf{85.96} & \textbf{85.98} & \textbf{86.27} & \textbf{86.01} \\ \hline
\end{tabular}
\label{tab4}
\end{table*}

\begin{table}[h!]
\centering
\caption{Confusion matrix. Fakeddit (2-way): Fake vs. Real; Fakeddit (3-way): Fake with true text; Fake with false text; Real; MFND (3-way): Fake, Real, Uncertain; Best classifications in bold.}
\begin{tabular}{|c|c|c|c|c|c|c|c|}
\hline
\multicolumn{2}{|c|}{\textbf{Fakeddit (2-way)}} & \multicolumn{3}{c|}{\textbf{Fakeddit (3-way)}} & \multicolumn{3}{c|}{\textbf{MFND (3-way)}} \\ \hline
415 & 3 & 200 & 2 & 6 & 174 & 7 & 2 \\ \hline
5 & 203 & 0 & 209 & 3 & 14 & 169 & 22 \\ \hline
\multicolumn{2}{|c|}{} & 4 & 1 & 202 & 14 & 24 & 165 \\ \hline
\end{tabular}
\label{tab5}
\end{table}
\subsection{Performance scores for MAGIC}
\subsubsection{Baselines Comparison}
By comparing with baseline models, we can also identify certain patterns. For instance, when considering only the performance of baseline models, text-based methods generally outperform visual-based methods. This phenomenon holds true for both Chinese and English datasets. This observation underscores the importance of text and visual modalities in the field of fake news detection. Importantly, it highlights that text, with its richer discriminative attributes, is a more favorable factor for enhancing detection capabilities.

In the realm of pure text-based baseline methods, especially on English datasets, the performance of almost all models seems quite comparable. In Fakeddit(2-way), Bert+CNN outperforms other baseline models, while when Fakeddit is categorized into more detailed classes, the performance of BERT+LSTM surpasses Bert+CNN. However, on Chinese datasets, the BERT-GNN method significantly outperforms other baseline models, indicating higher quality comment data on Chinese datasets, providing reliable supplementary information for fake news identification.

\subsubsection{Comparison between Baselines and MAGIC}
A quick overview of the results shows MAGIC to outperform the rest of the models, regardless of the datasets used with accuracy scores ranging from 85.96\% to 98.72\%, and F-scores ranging from 86.01\% to 98.56\%. This shows the effectiveness in classifying fake news using graph-based structures for multimodal input, regardless of the languages used. A finer comparison between input types revealed MAGIC to outperform all the single modality models, regardless of the datasets used. This shows that incorporating features from more than a single input type improves fake news classification, concurring with previous studies \cite{1,6,28,30}. 

A comparison of the datasets revealed that MAGIC produced significantly superior classification results for Fakeddit in contrast to MFND, including when compared against similar number of classifications (i.e., Accuracy$_\mathrm{Fakeddit}=97.60\%$ vs. $\mathrm{Accuracy}_{\text{MFND}19}=85.96\%).$ BERT supports multi-lingual texts including Chinese, however, it was originally trained and built based on English corpora, and thus probably leads to better text understanding and detection when textual input is provided in English. In contrast, the Chinese version may face challenges related to dialects, character variations, and language nuances \cite{37}. Nevertheless, MAGIC's performance on MFND is still undeniably good as indicated by the accuracy and F-scores ranging from 85.96\%- 86\%. Although a different approach and datasets were used, a similar finding was reflected whereby the authors found a multimodal fusion network to yield a better accuracy score on $\text{Politifact compared to}$ Weibo \cite{3}.

Looking at Fakeddit, MAGIC performed consistently well for the 2-way and 3-way setup, with accuracy scores ranging from 97.60\% to 98.72\%, and F-score ranging from 97.60\% to 98.56\%. These are in line with several fake news detection studies that have reported consistent performance scores between 2- and 3-way classification datasets, regardless of the methods adopted \cite{38}. Further, the advantages of incorporating a graph-based mechanism to classify fake news are obvious when the results of BERT and BERT-GNN are compared, in which marginal improvements can be noted across both the datasets. Specifically, BERT-GNN performed well in both the 2-way and 3-way setups, with accuracy scores improving by 1.12\% and 0.16\%, respectively compared to BERT. Similarly, an improvement of 6.09\% in accuracy and 6.36\% in F-score was observed for the MFND dataset.
\subsection{Performance scores for the ablation study}
\begin{figure}[htp]
    \centering
    \includegraphics[width=1\linewidth]{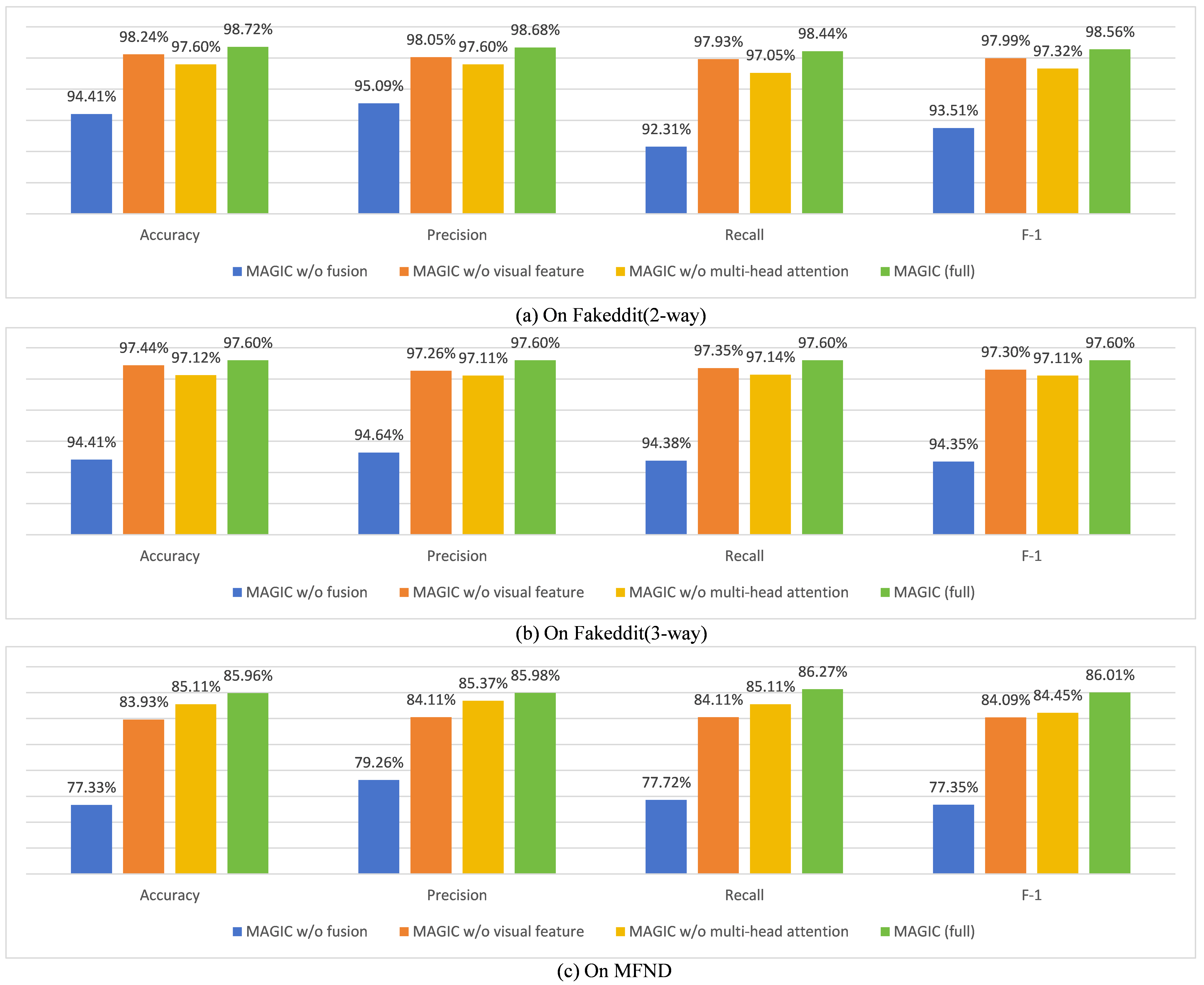}
    \caption{Results of ablation study on two datasets (in \%)}
    \label{fig:7}
\end{figure}

To further assess MAGIC’s performance, an ablation study was performed by systematically decomposing the model into simplified variants. This meticulous analysis aimed to identify and understand the specific components or features within the MAGIC model that contribute significantly to its overall effectivenes. Specifically, we (i) used our MAGIC model, but removes the visual-feature (excluded images), and (ii) omitted the multi-head attention mechanism (i.e., replacing GAN module in Figure \ref{fig:6} with a regular GNN). Performance scores in Figure \ref{fig:7} illustrates MAGIC to consistently outperform the rest of the models, albeit insignificantly in some of the instances. This shows the importance of each of the modules/layers added into our proposed graph-based fake news classification model.

It has been observed that the direct removal of the fusion layer results in a direct decrease in performance (i.e., MAGIC without fusion). The exclusion of image content from MAGIC showed a decline in classification performance for accuracy, precision, recall and F scores, that is, a decrease by 0.48\%, 0.63\%, 0.51\%, and 0.57\%, respectively compared to MAGIC on the Fakeddit 2-way dataset. Similar declining patterns were observed for the 3-way Fakeddit and MFND datasets. This demonstrates the importance of simultaneously integrating modal and structural features for fake news classification \cite{21}. Similar observations were also noted for the other models, hence indicating images and their features to contributes to improving fake news classification, a finding that well concurs with previous studies such as \cite{31}. Findings in Figure \ref{fig:7} affirms the importance of integrating multimodal input (i.e., text and images) incorporating the multi-head attention mechanism that helps to enhance the stability and performance of fake news detection models, regardless of the languages \cite{16}.

\section{Conclusion}\label{s6}
This study proposed a Multimodal Adaptive Graph-based Intelligent
Classification Model, 
MAGIC, that was trained and tested on two datasets (English and Chinese). Original posts, 
their accompanying comments and images were incorporated as part of MAGIC, hence 
enabling the formation of interaction graphs. MAGIC not only simplifies the intricate 
process of graph creation but also accomplishes dual objectives within a single network: 
modal fusion and capturing comment structures. Experiments in various scenarios 
including the ablation experiments revealed MAGIC to consistently outperform the 
baseline models.
\subsection{Limitations}
The study is not without its limitations: first, it relies on extracting features from different modalities through embedding. Therefore, the quality of the embedding model will have a significant impact on the results. Future studies could explore other embedding techniques to improve the quality of feature extraction based on the diverse modalities. Second, although the pre-trained transfer learning models were adapted, future studies could replicate MAGIC on larger datasets and expand its applicability to other languages, including low-resourced ones, given the extensive and diverse linguistic content present on social media, spanning various languages, nuances, and dialects. Finally, MAGIC was developed to handle text and images only, therefore the model can be expanded to handle videos and audios using other multimodal fusion techniques.
\subsection{Future Work}
Fake news detection is a field that requires continuous iteration of data processing techniques and models, leaving ample room for future research. 

For instance, in our current research, the datasets include only posts, images, and comments, predominantly sourced from Twitter and Weibo. However, emerging social media platforms with enhanced dissemination capabilities, such as TikTok, have come into play. TikTok boasts robust recommendation algorithms, facilitating not only faster but also more precise targeting of user demographics. The mode of information propagation on TikTok extends beyond mere text and images, encompassing richer multimodal content like videos and audio. Furthermore, its recommendation algorithms are intricately tied to users' profiles, behaviors, and the entirety of their social networks, presenting new challenges in the detection of fake news. 

Additionally, with the skyrocketing popularity of generative large language models like ChatGPT, an increasing amount of information is becoming more susceptible to manipulation and fabrication. Many instances of false information are directly generated by large language models (LLMs), while manipulated images are produced by image generation models like midjourney. This poses additional challenges to current efforts in combating fake news. 

Therefore, future work can be expanded in the following directions: (i) collecting fake news datasets that encompass a more diverse range of modalities; (ii) exploring and developing more robust and versatile detection models capable of handling different modalities; (iii) incorporating the generative capabilities of large pre-trained language models (LLMs) to extend the fake news classification task into an end-to-end debunking system. This system not only discerns the veracity of fake news but also provides users with content generated by LLMs to enhance the interpretability of the information. For instance, for verified information, the system could provide the source, for fake information, it could offer refutations, and for uncertain information, it could guide users toward further investigation.

\bibliographystyle{IEEEtran}
\bibliography{reference}

\end{document}